\documentclass{article}

\usepackage{arxiv}

\usepackage[utf8]{inputenc} 
\usepackage[T1]{fontenc}    
\usepackage{hyperref}       
\usepackage{url}            
\usepackage{booktabs}       
\usepackage{amsfonts}       
\usepackage{nicefrac}       
\usepackage{microtype}      
\usepackage{lipsum}
\usepackage{graphicx}

\usepackage{subfigure}
\usepackage{diagbox}
\usepackage{makecell}
\usepackage{tabularx}

\graphicspath{ {./img/} }

\title{Lookup subnet based Spatial Graph Convolutional neural Network}

\author{
 HU Jingzhao \\
  Department of Information Science and Technology \\
  Northwest University \\
  Xi’an 710127, China \\
 \And
 ZHANG Xiaoqi \\
  Department of Information Science and Technology \\
  Northwest University \\
  Xi’an 710127, China \\
 \And
 JIA Qiaomei \\
  Department of Information Science and Technology \\
  Northwest University \\
  Xi’an 710127, China \\
  \And
 WANG Chen \\
  Department of Information Science and Technology \\
  Northwest University \\
  Xi’an 710127, China \\
  \And
 BU Qirong \\
  Department of Information Science and Technology \\
  Northwest University \\
  Xi’an 710127, China \\
  \And
 FENG Jun$^*$ \\
  Department of Information Science and Technology \\
  Northwest University \\
  Xi’an 710127, China \\
  \texttt{fengjun@nwu.edu.cn}\\
}

\begin{document}
\maketitle
\begin{abstract}
	Convolutional Neural Networks(CNNs) has achieved remarkable performance breakthrough in Euclidean structure data. Recently, aggregation-transformation based Graph Neural networks(GNNs) gradually produce a powerful performance on non-Euclidean data. In this paper, we propose a cross-correlation based graph convolution method allowing to naturally generalize CNNs to non-Euclidean domains and inherit the excellent natures of CNNs, such as local filters, parameter sharing, flexible receptive field, etc. Meanwhile, it leverages dynamically generated convolution kernel and cross-correlation operators to address the shortcomings of prior methods based on aggregation-transformation or their approximations. Our method has achieved or matched popular state-of-the-art results across three established graph benchmarks: the Cora, Citeseer, and Pubmed citation network datasets. 
\end{abstract}

\keywords{deep learning \and graph neural network \and graph convolutional neural network \and lookup subnet based spatial graph convolutional neural network}

	\section{Introduction}
		Convolutional Neural Networks(CNNs) have been successfully applied to tasks that the underlying data representation has a Euclidean structure, such as image classification\cite{7780459}, semantic segmentation\cite{J2017The}, and machine translation\cite{MachineTranslation}, etc. However, the data for lots of tasks can't be represented as a Euclidean structure, such as social network data, telecommunication network data, paper citation data, and brain connectome data, etc. Such non-Euclidean data are usually represented in the form of graphs.
		
		There have been several works attempting to use neural networks to process arbitrarily structured graphs. Due to the data struct, recursive neural networks were earliest employed to process data represented in graph domains\cite{Frasconi1998A}\cite{Sperduti1997Supervised}\cite{Gori2005A}\cite{Scarselli2009The}. The idea was adopted and improved to use gated recurrent units\cite{Cho2014Learning} by \cite{Li2015Gated}. Recently, there is an increasing interest in generalizing convolutions to the graph domain. Such works are often categorized as spectral approaches and spatial approaches. The spectral approaches are more mathematical driven. They first transform the graphs to the spectral domain by graph Fourier transform, then define convolution kernels in the spectral domain, then perform convolution in the spectral domain, and finally invert back to the graph domain\cite{Bruna2013Spectral}\cite{Henaff2015Deep}\cite{Defferrard2016Convolutional}\cite{Kipf2016Semi}\cite{wang2020noise}. The spatial approaches directly define convolutions on graphs, operating on spatially close neighbors. The biggest challenges of spatial approaches are to filter on different sized neighborhoods. To address this, some works require learning a specific weight matrix for each node degree\cite{Duvenaud2015Convolutional} , some works require extracting and normalizing neighborhoods containing a fixed number of nodes\cite{Niepert2016Learning}. Mixture model CNNs (MoNet) presented by \cite{Monti2016Geometric} provides a unified generalization of CNN architectures to graphs. GraphSAGE introduced by \cite{Hamilton2017Inductive} operates by sampling neighborhood of each node, and then performing a specific aggregator over it. 
		
		Almost all of the works mentioned above can be described as aggregation-transformation based methods, which means that they are used to perform feature extraction of graphs through a set of aggregation operators and transformation operators. The aggregation operator nicely solves the problem of filtering on different sized neighborhoods, but it ignores that the neighbors may have different importance for feature extraction. There have been several works that work on assigning different weights to different neighbors for aggregation operators, such as MoNet\cite{Monti2016Geometric} based on distance metrics, GAT\cite{Veli2017Graph} based on attention, GIN\cite{Xu2018How}, etc. However,  the differences in the importance of the channel dimensions haven't been aware.
		
		Inspired by the shortcomings of aggregation-transformation based GNNs and the reputation of CNNs, we introduce a cross-correlation based graph convolution method, which we reference as Lookup subnet based Spatial Graph Convolutional neural Network(LSGCN), allowing to naturally generalize CNNs to non-Euclidean domains and inherit the excellent nature of CNNs. The idea is to dynamically generate convolution kernels by looking up weights in a sub neural network using auto-encoded node position code. The introduced architecture has several interesting properties:(1)the cross-correlation operator simultaneously considers the difference in the importance of neighbors and channel dimensions; (2)flexible receptive field definition; (3)the layer capacity flexibility adjust by kernel size (lookup subnet) and filter number; (4)support in-graph mini-batch allowing to perform on large graphs; and (5) local filters and parameter sharing, which directly inherit from CNNs. We validate the proposed approach on three challenging benchmarks: Cora, Citeseer and Pubmed citation networks, achieving or matching popular state-of-the-art results that highlight the potential of cross-correlation based spatial models when dealing with arbitrarily structured graphs.
		
	\section{Related Work}
		In the early times, recursive neural networks were employed to process data represented in graph domains as a directed acyclic graph\cite{Frasconi1998A}\cite{Sperduti1997Supervised}. \cite{Gori2005A} and \cite{Scarselli2009The} generalized recursive neural networks to directly deal with cyclic, directed and undirected graphs. Later, the idea was adopted and improved to use gated recurrent units\cite{Cho2014Learning} by \cite{Li2015Gated}. 
		
		Convolutional neural networks(CNNs) have achieved impressive performance on various tasks, and many researchers attribute its powerful performance to its local filters and parameter sharing properties. Recently, there is an increasing interest in generalizing convolutions to the graph domain. Such works are often categorized as spectral approaches and spatial approaches. 
		
		The spectral approaches are more mathematical driven. It first transforms the graphs to the spectral domain by graph Fourier transform, then defines convolution kernels in the spectral domain, then performs convolution in the spectral domain, and finally inverts back to the graph domain\cite{Bruna2013Spectral}. Later, \cite{Henaff2015Deep} improved spectral filters with smooth coefficients in order to make them spatially localized. \cite{Defferrard2016Convolutional} proposed to approximate the filters by Chebyshev expansion, removing the need to reduce the amount of calculation. Finally, \cite{Kipf2016Semi} simplified the previous method by restricting the filters to operate in a 1-step neighborhood.
		
		The spatial approaches directly define convolutions on graphs, operating on spatially close neighbors. The biggest challenges of spatial approaches are to filter on different sized neighborhoods. To address this, some works require learning a specific weight matrix for each node degree\cite{Duvenaud2015Convolutional} , some works require extracting and normalizing neighborhoods containing a fixed number of nodes\cite{Niepert2016Learning}. Mixture model CNNs (MoNet) presented by \cite{Monti2016Geometric} provides a unified generalization of CNN architectures to graphs. GraphSAGE introduced by \cite{Hamilton2017Inductive} operates by sampling neighborhood of each node, and then performing a specific aggregator over it.
		
		Lately, researchers have noted that the neighbors may have different importance for feature extraction, and there have been several works that work on assigning different weights to different neighbors, such as MoNet\cite{Monti2016Geometric} based on distance metrics, GAT\cite{Veli2017Graph} based on attention, GIN\cite{Xu2018How}, etc.

	\section{Methodology}
		In this section, we will firstly present the building block layer used to construct our graph convolutional neural networks, which we reference as Lookup subnet based Spatial Graph Convolutional layer(LSGC layer). Then, we will introduce a graph convolutional neural network constructed through stacking LSGC layer, which we reference as Lookup subnet based Spatial Graph Convolutional neural Network(LSGCN).
		
		\subsection{Lookup subnet based Spatial Graph Convolutional layer}
			The motivation is to dynamically generate convolution kernels by looking up weights in a sub neural network using auto-encoded node position code. This motivation brings the possibility of natural generalizing CNNs to non-Euclidean domains and inheriting the excellent nature of CNNs.
		
			We consider a multi-layer Graph Convolutional Network constructed through stacking Lookup subnet based Spatial Graph Convolutional layer(LSGC layer) with the following defined layer-wise propagation rule:
			
			\begin{equation}
				\label{equ:lsgc}
				H^{(l+1)}(i, k) = \delta(\sum{\{L_k(E(P_i) - E(P_{\{N_A(i, r)\}})) \otimes H^{(l)}_{\{N_A(i,r)\}}}\} / |\{N_A(i,r)\}| + b_k)
			\end{equation}
			
			where $H^{(l)}$ is the matrix of activations in the $l^{th}$ layer. $P$ is node position indices that randomly assigned to nodes. $A$ is the the adjacency matrix of the undirected graph with self-connections. $b$ is the bias. $\delta(\cdot)$ denotes an activation function; $L$ is the dynamical convolution kernel generator defined by a neural network, which we reference as lookup subnet. $E$ is the node position encoder defined by a neural network that starts with an embedding layer. $N_A$ is a function that calculates the neighborhoods of previously defined adjacency matrix $A$.
			
			Further, $H^{l}(i,k)$ is the activation of $k^{th}$ kernel of $i^{th}$ node. $H^{l}_{\{N_A(i, r)\}}$ is the matrix of activations in $r$-order neighborhoods of $i^{th}$ node, where $r$ represents the receptive field. $L_k$ is the lookup subnet of $k^{th}$ kernel. $P_i$ is the node position index of $i^{th}$ node. $P_{\{N_A(i, r)\}}$ is the node position indices of $r$-order neighborhoods of $i^{th}$ node. $|\{N_A(i,r)\}|$ is the size of $r$-order neighborhoods of $i^{th}$ node. $b_k$ is bias of $k^{th}$ kernel.
			
			Steply, assume we would extract features for $i^{th}$ node in a graph. We first obtain neighbors in the receptive field of convolutional center node: $N_A(i, r)$. Then, we calculate the auto-encoded node position code of neighbors: $E(P_{\{N_A(i, r)\}})$. Then, we fuse the auto-encoded node position code of convolutional center node and it's neighbors: $E(P_i) - E(P_{\{N_A(i, r)\}})$. Finally, we generate kernels for the convolutional center by $L$ using fused node position code and perform cross-correlation operator, i.e. Equation \ref{equ:lsgc}.
			
			\begin{figure}[htbp]
				\centerline{
					\subfigure{\includegraphics[width=0.35\textwidth]{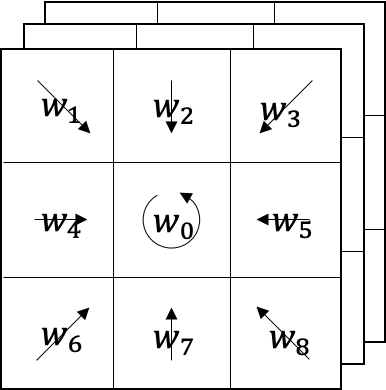}}
					\\
					\subfigure{\includegraphics[width=0.35\textwidth]{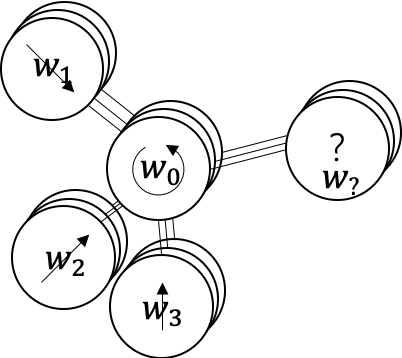}}
				}
				\caption{\textbf{Left:} The convolutional layer extracts features by cross-correlation operator, parametrized by its kernels $W = \{w_0, \dots, w_8\}$ defined directly. \textbf{Right:} The lsgc layer also extracts features by cross-correlation operator, parametrized by the parameters that used to parameterize lookup subnets. Its kernels $w = \{w_0, \dots, w_?\}$ dynamically generate by looking up weights in lookup subnets using auto-encoded node position codes. Where $?$ denotes the uncertain node reference to different sized neighborhoods.}
				\label{fig:cov and lsgc}
			\end{figure}
			
			The similarities and differences of basic principles between the convolutional layer and lsgc layer are illustrated by Figure \ref{fig:cov and lsgc}.
			

		\subsection{lookup subnet based spatial graph convolutional neural network}
			In this subsection, we introduce a neural network architecture mainly constructed by stacking lsgc layers to perform node classification of graph-structured data, which we call Lookup subnet based Spatial Graph Convolutional neural Network, reference as LSGCN.
			
			\begin{figure}[htbp]
				\centering{\includegraphics[width=0.95\textwidth]{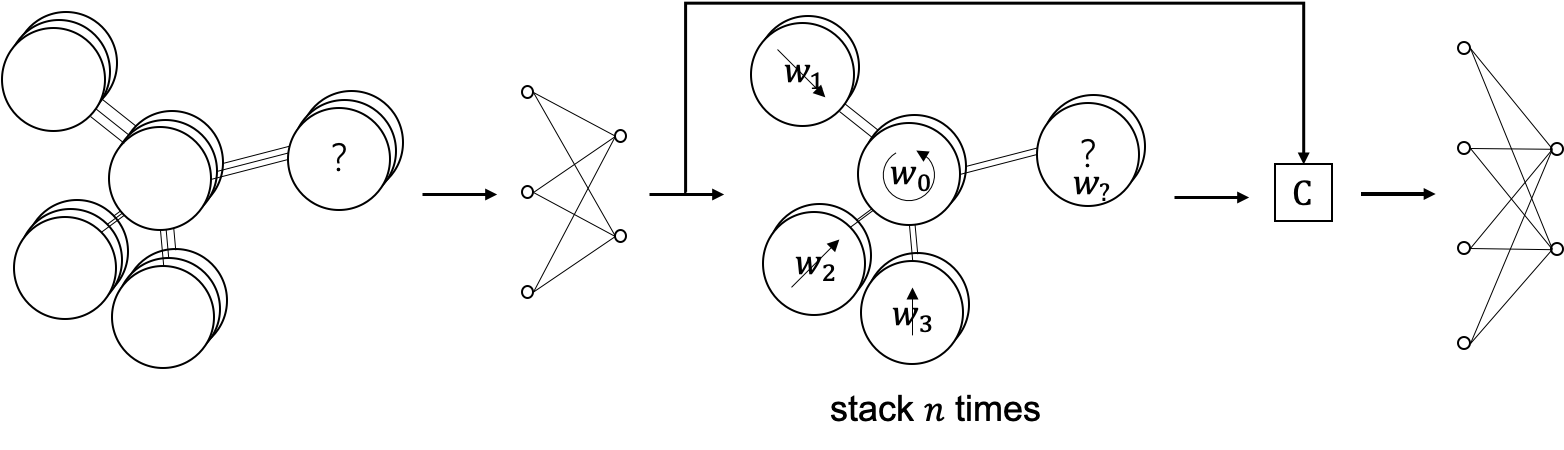}}
				\caption{Lookup subnet based Spatial Graph Convolutional neural Network, reference as LSGCN. Node features are first transformed by fully connected layers. Then, the graph with transformed node features is feed to lsgc layers. Finally, the extracted node feature is concatenated with transformed node features, and these concatenate features are feed to a fully connected layer to perform classification. Where $C$ denotes the feature concatenate operator. $n$ is the number of stacked lsgc layers.}
				\label{fig:lsgcn}
			\end{figure}
			
			The LSGCN architecture is illustrated in Figure \ref{fig:lsgcn}. Considering that the lsgc layer that mainly used to construct LSGCN is based on cross-correlation which simultaneously extracts features from spatial and channel dimensions, we especially illustrate the channel dimension by carefully stacking the illustration of graphs.
			
			The node features are first transformed by fully connected layers to adjust the dimension of features which decide the number of the input feature maps of following lsgc layers. Then, the graph with transformed node features is feed to lsgc layers to extract higher-level features from its struct. Finally, the extracted node feature is concatenated with transformed node features, and these concatenate features are feed to a fully connected layer to perform classification. It is worth noting that, the architecture has various flexible ways to adjust the model capacity, we should carefully determine its instances. 
		
	\section{Experimental \& Results}
		We have performed a comparative evaluation of LSGCN against a wide variety of strong baselines and previous approaches, on three established graph-based benchmark tasks, achieving or matching popular state-of-the-art performance across all of them. In this section, we first introduce the datasets, then summarize the experimental setups, finally report the results. 
				
		\subsection{Datasets}
			Three standard citation network benchmark datasets are utilized to perform the comprehensive evaluations: Cora, Citeseer, and Pubmed\cite{Sen2008Collective}. We closely follow the dataset setups of \cite{Veli2017Graph}. 
			
			In all of the used datasets, nodes correspond to documents and undirected edges to citations. Node features correspond to elements of a bag-of-words representation of a document. Each node has a class label. Following the dataset setups from previous studies, we allow for only 20 nodes per class to be used for training. The predictive performance of the trained models is evaluated on 1000 test nodes. These training and test sets are the same ones as used by \cite{Veli2017Graph}. The Cora dataset contains 2708 nodes, 5429 edges, 7 classes, and 1433 features per node. The Citeseer dataset contains 3327 nodes, 4732 edges, 6 classes, and 3703 features per node. The Pubmed dataset contains 19717 nodes, 44338 edges, 3 classes, and 500 features per node. An overview of the interesting attributes of the datasets is given in Table \ref{tab:datasets}.
			
			\begin{table}[htbp]
				\centering
				\caption{Summary of the datasets used in our experiments.}
				\begin{tabularx}{0.95\textwidth}{c | >{\centering\arraybackslash}X >{\centering\arraybackslash}X >{\centering\arraybackslash}X}
					\toprule
					\diagbox{Attributes}{Datasets} & Cora & Citeseer & Pubmed \\
					\midrule
					Nodes & 2708 & 3327 & 19717 \\
					Edges & 5429 & 4732 & 44338 \\
					Features/Node & 1433 & 3703 & 500 \\
					Classes & 7 & 6 & 3 \\
					Training Nodes & 140 & 120 & 60 \\
					Test Nodes & 1000 & 1000 & 1000 \\
					\bottomrule
				\end{tabularx}
				\label{tab:datasets}
			\end{table}
			
		\subsection{Experimental setup}
			For the instance used to perform the evaluations of LSGCN architecture, all lsgc layers follow a same sub-design. The lookup subnets $L$ are designed as a struct of three fully connected layers. The node position encoders $E$ are implemented as a struct of a single embedding layer. The node position indices $P$ are assigned from the node's index itself. The activation functions $\delta(\cdot)$ are chosen as $\tanh(\cdot)$. 
			
			Its hyperparameters are automatically tuned using the BOHB\cite{Falkner2018BOHB} algorithm. The overview of tuned hyperparameters is shown in Table \ref{tab:hyperparameters}. Where the transformed node feature dimension is the output dimension of the fully connected layers in Figure \ref{fig:lsgcn}. The auto-encoded node position code dimension is the output dimension of the node position encoder $E$ in Equation \ref{equ:lsgc}. The number of kernels and receptive field are the hyperparameters for lsgc layers. In-graph mini-batch size denotes the number of nodes that fed to LSGCN for one updating to the model. 
			
			\begin{table}[htbp]
				\centering
				\caption{Overview of hyperparameters automatically tuned using the BOHB\cite{Falkner2018BOHB} algorithm.}
				\begin{tabularx}{0.95\textwidth}{c | >{\centering\arraybackslash}X >{\centering\arraybackslash}X >{\centering\arraybackslash}X}
					\toprule
					\diagbox{Hyperparameters}{Datasets} & Cora & Citeseer & Pubmed \\
					\midrule
					\makecell[c]{Transformed node feature \\ dimension} & 355 & 1401 & 502 \\
					\makecell[c]{Auto-encoded node position \\ code dimension} & 234 & 134 & 6 \\
					Number of kernels & 185 & 205 & 68 \\
					Receptive field & 2 & 3 & 3 \\
					Learning rate & 0.0020 & 0.0002 & 0.0029 \\
					In-graph mini-batch size & 8 & 8 & 8 \\
					\bottomrule
				\end{tabularx}
				\label{tab:hyperparameters}
			\end{table}
			
			All experiments in this paper were conducted using a Geforce RTX 2080 Ti. The machine learning framework used in this paper is PyTorch\cite{pytorch}. The hyperparameter tuning framework used in this paper is NNI\cite{nni}.
		
		\subsection{Results}
			The results of our comparative evaluation experiments are summarized in Table \ref{tab:results}. We report the mean classification accuracy on the test nodes of our method, and reuse the metrics already reported in \cite{Veli2017Graph} and \cite{Thekumparampil2018Attention} for popular state-of-the-art techniques.
			
			\begin{table}[htbp]
				\centering
				\caption{Summary of results in terms of classification accuracies.}
				\begin{tabularx}{0.95\textwidth}{c | >{\centering\arraybackslash}X >{\centering\arraybackslash}X >{\centering\arraybackslash}X}
					\toprule
					\diagbox{Method}{Accuracies}{Datasets} & Cora & Citeseer & Pubmed \\
					\midrule
					MLP & 55.1\% & 46.5\% & 71.4\% \\
					ManiReg\cite{Belkin2004Manifold} & 59.5\% & 60.1\% & 70.7\% \\
					SemiEmb\cite{Weston08deeplearning} & 59.0\% & 59.6\% & 71.7\% \\
					LP\cite{Zhu2003Semi} & 68.0\% & 45.3\% & 63.0\% \\
					DeepWalk\cite{10.1145/2623330.2623732} & 67.2\% & 43.2\% & 65.3\% \\
					ICA\cite{Getoor2005Link} & 75.1\% & 69.1\% & 73.9\% \\
					Planetoid\cite{10.5555/3045390.3045396} & 75.7\% & 64.7\% & 77.2\% \\
					Chebyshev\cite{Defferrard2016Convolutional} & 81.2\% & 69.8\% & 74.4\% \\
					GCN\cite{Kipf2016Semi} & 81.5\% & 70.3\% & 79.0\% \\
					GCN-64\cite{Veli2017Graph} & 81.4\% & 70.9\% & 79.0\% \\
					MoNet\cite{Monti2016Geometric} & 81.7\% & - & 78.8\% \\
					AGNN\cite{Thekumparampil2018Attention} & 82.6\% & 71.7\% & 79.9\% \\
					GAT\cite{Veli2017Graph} & 83.0\% & 72.5\% & 79.0\% \\
					\midrule
					\textbf{LSGCN}(ours) & \textbf{83.3}\% & \textbf{73.0}\% & \textbf{80.0}\% \\
					\bottomrule
				\end{tabularx}
				\label{tab:results}
			\end{table}
			
			Our results successfully demonstrate the popular state-of-the-art performance being achieved or matched across all three datasets in concordance with our expectations. The results demonstrate the feasibility and significance of the cross-correlation based lsgc method.
			
	\section{Discussion}
		Considering that the hyperparameters in Table \ref{tab:hyperparameters} are automatically optimized by the tuning algorithm, these optimized hyperparameters are more dependent on the characteristics of the LSGCN itself. This in turn eliminates the author's preference in setting the hyperparameters. Thus we can use these optimized hyperparameters to reverse the characteristics that the LSGCN has.
		
		The most interesting hyperparameters are the number of kernels and the receptive field, they exhibit similar patterns to the number of kernels and the kernel size in CNNs. The number of kernels determines the feature dimension of activation in lsgc layer, since the cross-correlation based lsgc layer simultaneously extracts features from spatial and channel dimensions. The receptive fields show that it is beneficial for extracting features, although in most prior methods based on aggregation-transformation or their approximations this value is usually fixed to 1. And the in-graph mini-batch size, it has a real impact on the learning process, not just determine the required resources for a batch. All concepts inherited from CNNs behave as they just work in CNNs, which is in concordance with our expectations.
				
	\section{Conclusion}
		We have presented Lookup subnet based Spatial Graph Convolutional layer(LSGC layer) and Lookup subnet based Spatial Graph Convolutional neural Network(LSGC), novel cross-correlation based graph convolutional layer and neural network that operate on non-Euclidean data, leveraging dynamically generated convolution kernels by looking up weights in a sub neural network using auto-encoded node position code. We demonstrate that it can naturally generalize CNNs to non-Euclidean domains and successfully inherit the excellent natures and concepts of CNNs. Thus it addresses many shortcomings of prior methods based on aggregation-transformation or their approximations. Our LSGCN models leveraging LSGC layers have successfully achieved or matched popular state-of-the-art performance across three well-established node classification benchmarks.
		
		
	\section*{Acknowledgements}
		This work was supported by the National Key Research and Development Program of China under grant 2017YFB1002504.

\bibliographystyle{unsrt}  
\bibliography{references}  

\end{document}